\documentclass[sigconf]{acmart}
\renewcommand\footnotetextcopyrightpermission[1]{} 
\usepackage{multirow}
\usepackage{minibox}
\usepackage[ruled,vlined]{algorithm2e}
\DontPrintSemicolon
\usepackage{listings}
\usepackage{adjustbox}

 \newcommand\FramedBox[3]{%
   \setlength{\fboxrule}{0pt}
   \fbox{\parbox[t][#1][c]{#2}{\raggedright\small #3}}}

\usepackage{xcolor}

\usepackage{booktabs} 
\usepackage{subcaption}
\usepackage{enumitem}
\usepackage{footnote}
\usepackage{multirow}

\usepackage{balance}

\setcopyright{none}

\begin{document}
\title[Product Characterisation towards Personalisation]{Product Characterisation towards Personalisation}
\subtitle{Learning Attributes from Unstructured Data to Recommend Fashion Products}

\author{\^Angelo Cardoso}
\authornote{work developed while at ASOS.com}
\orcid{0000-0003-4155-3827}
\affiliation{%
	\institution{ISR, IST, Universidade de Lisboa}
	\streetaddress{Av. Rovisco Pais}
	\city{Lisbon} 
    \state{Portugal} 
	\postcode{NW1 7FB}
}
\email{angelo.cardoso@tecnico.ulisboa.pt}

\author{Fabio Daolio}
\orcid{0000-0003-4240-4161}
\affiliation{%
	\institution{ASOS.com}
	\streetaddress{Greater London House,
Hampstead Road}
	\city{London} 
    \state{UK} 
	\postcode{NW1 7FB}
}
\email{fabio.daolio@asos.com}

\author{Sa\'ul Vargas}
\orcid{0000-0002-3459-3102}
\affiliation{%
	\institution{ASOS.com}
	\streetaddress{Greater London House,
Hampstead Road}
	\city{London} 
    \state{UK} 
	\postcode{NW1 7FB}
}
\email{saul.vargassandoval@asos.com}

\renewcommand{\shortauthors}{\^A. Cardoso et al.}

\begin{abstract}
In this paper, we describe a solution to tackle a common set of challenges in e-commerce, which arise from the fact that new products are continually being added to the catalogue. The challenges involve properly personalising the customer experience, forecasting demand and planning the product range. We argue that the foundational piece to solve all of these problems is having consistent and detailed information about each product, information that is rarely available or consistent given the multitude of suppliers and types of products. We describe in detail the architecture and methodology implemented at ASOS, one of the world's largest fashion e-commerce retailers, to tackle this problem. We then show how this quantitative understanding of the products can be leveraged to improve recommendations in a  hybrid recommender system approach.
\end{abstract}



\maketitle

\section{Introduction}
ASOS is a global e-commerce company that creates and curates clothing and beauty products for fashion-loving 20-somethings. All of the company's sales are originated online via mobile apps and country-specific websites. 
ASOS's websites attracted 174 million visits during December 2017 (December 2016: 139 million) and as at 31 December 2017 it had 16.0 million active customers.

At any moment in time, ASOS's catalogue can offer around 85K products to its customers, with around 5K new propositions being introduced every week across its platforms. Over the years, this amounts to more than one million unique styles, that is, without accounting for size variants. For each of these products, different divisions within the company produce and consume different product attributes, most of which are not even customer-facing and, hence, are not always present and consistent. However, these incomplete attributes still carry information that is relevant to the business. 

The ability to have a systematic and quantitative characterisation of each of these products, particularly the yet-to-be-released ones, is key for the company to make data-driven decisions across a set of problems including personalisation, demand forecasting, range planning and logistics. In this paper, we show how we predict a coherent and complete set of product attributes and illustrate how this enables us to personalise the customer experience by providing more relevant products to customers.

\begin{figure*}[h]
\centering
\subcaptionbox{Input images\label{img}}{%
	\begin{tabular}{c@{}c}
	\includegraphics[width=0.105\textwidth]{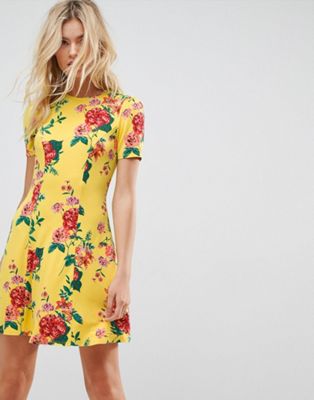}
    &\includegraphics[width=0.105\textwidth]{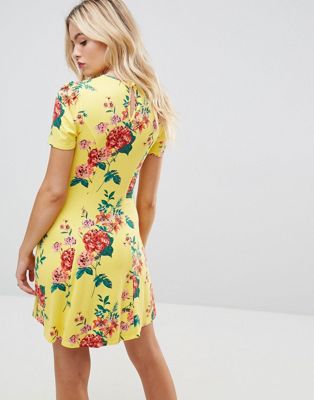}\\[-0.12cm]
    \includegraphics[width=0.105\textwidth]{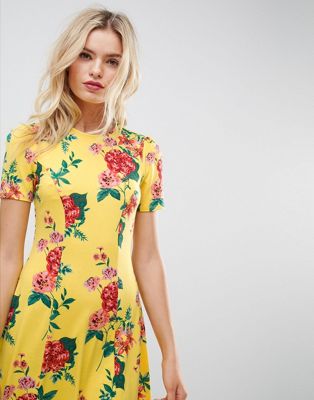}
    &\includegraphics[width=0.105\textwidth]{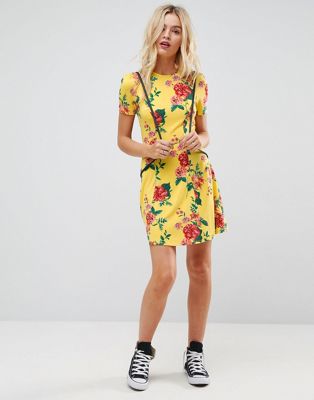}\\[0.01cm]
	\includegraphics[width=0.105\textwidth]{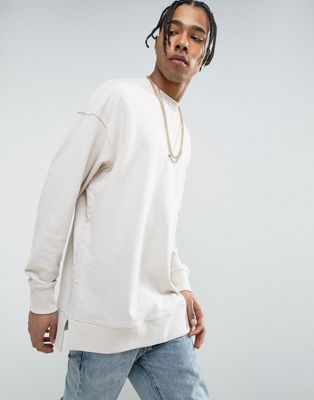}
    &\includegraphics[width=0.105\textwidth]{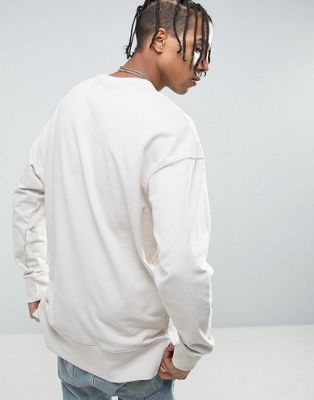}\\[-0.12cm]
    \includegraphics[width=0.105\textwidth]{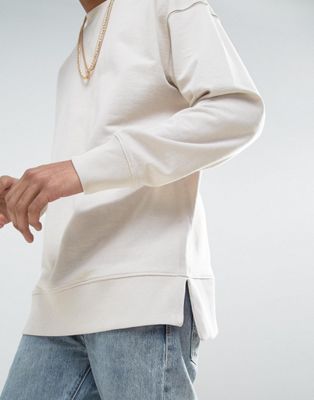}
    &\includegraphics[width=0.105\textwidth]{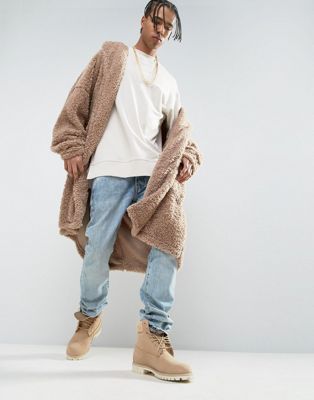}\\[-0.1cm]
	\end{tabular}}
\subcaptionbox{Input text\label{txt}}{%
	\begin{tabular}{c}
	\FramedBox{0.25\textwidth}{0.31\textwidth}{\textbf{ASOS Mini Tea Dress In Yellow Floral Print}\\
    \noindent Dress by ASOS Collection\\
    \begin{itemize}[leftmargin=*]
    \item Floral print fabric
    \item Crew neck
    \item Button-keyhole back
    \item Fit-and-flare style
    \item Regular fit - true to size
    \item Machine wash
    \item 94\% Viscose, 6\% Elastane
    \item Our model wears a UK 8/EU 36/US 4 and is 176 cm/5'9.5" tall
    \end{itemize}}\\
	\FramedBox{0.25\textwidth}{0.31\textwidth}{\textbf{New Look Oversized Dropped Shoulder Sweatshirt In Ecru}\\
    \noindent Sweatshirt by New Look\\
    \begin{itemize}[leftmargin=*]
    \item Loop-back sweat
    \item Crew neck
    \item Dropped shoulders
    \item Ribbed trims
    \item Step hem
    \item Oversized fit - falls generously over the body
    \item 80\% Cotton, 20\% Polyester
    \item Our model wears a size Medium and is 188 cm/6'2" tall
    \end{itemize}}
    \end{tabular}}
\subcaptionbox{Attributes output\label{out}}{%
	\begin{tabular}{c}
	\FramedBox{0.25\textwidth}{0.31\textwidth}{
	\begin{description}[leftmargin=*]
	\item[DRESS TYPE] casual dresses
    \item[PATTERN] floral
    \item[BACK TO SCHOOL/COLLEGE] no
    \item[CUSTOMER SEGMENT] girly girl
    \item[USE/OCCASION] day casual
    \item[STYLE] tea dresses
    \item[SEASONAL EVENT] high summer
    \item[RANGE] main collection
    \item[PRICE RANGE] high street
    \item[CATEGORY] day dresses - jersey print
    \item[LENGTH] mini
    \item[...] 
	\end{description}}\\
	\FramedBox{0.25\textwidth}{0.31\textwidth}{
	\begin{description}[leftmargin=*]
	\item[PRODUCT LENGTH] longline
    \item[PATTERN] plain
    \item[BACK TO SCHOOL/COLLEGE] no
    \item[CUSTOMER SEGMENT] street lux
    \item[STYLE] sweatshirts
    \item[SEASONAL EVENT] cold weather
    \item[RANGE] main collection
    \item[PRICE RANGE] high street
    \item[PRODUCT FIT] oversized
    \item[NECKLINE] crew
    \item[SLEEVE LENGTH] long sleeve
    \item[...] 
	\end{description}}
    \end{tabular}}
\caption{Example instances with their associated (a) product images and (b) product details, as they appear on product pages. Using these inputs together with a product's type, brand and division, the network predicts a set of product attributes (c).}
\label{fig:probleminputoutput}
\end{figure*}

To achieve a coherent characterisation of all products, we leverage manually annotated attributes that only cover part of the product catalogue as illustrated in Figure~\ref{fig:statusquo}. As data that is consistently available for all products, we have a set of images, a text description as well as product type and brand, as shown in Figure~\ref{fig:probleminputoutput}.


Aiming to show how customer experience can be improved by enriching product/content data, this paper contributes:
\begin{enumerate}
\item the description of a real use case where augmenting product information enables better personalisation;
\item a system for consolidating product attributes that deals with missing labels in a multi-task setting and at scale;
\item a hybrid recommender system approach using vector composition of content-based and collaborative components.
\end{enumerate}

\noindent The key features of the product annotation pipeline are:
     \begin{itemize}
     	\item{shared model -- a neural network where most of layers are shared across the tasks (i.e. the attributes) to leverage relationship between the tasks as well as to reduce the total number of free parameters;}
        \item{training algorithm -- a custom model fitting procedure to achieve a balanced solution across all tasks and to deal with partially-labelled data.}
     \end{itemize}
The main aspects of the hybrid recommender system are:
     \begin{itemize}
     	\item{model composition -- the hybrid model is a vector composition of a collaborative and content-based models;}
        \item{simultaneous optimisation -- we optimise the two components of the hybrid model simultaneously.}
     \end{itemize}

The remainder of the paper is organised as follows: Section~\ref{sec:attr} presents the product attribute prediction pipeline and related literature, Section~\ref{sec:recs} illustrates how product information can improve recommendations and related literature, Section~\ref{sec:discussion} 
presents a discussion.

\begin{figure}[t]
\begin{tabular}{ccccc}
\toprule
product & type & segment & pattern &...\\
\midrule
A & dress &? &floral &?\\
B & dress &girly girl &? &...\\
C & skirt &? &check &?\\
... &... &... &... &...\\
\bottomrule
\end{tabular}
\caption{Illustrative representation of the problem with the original product data -- most products have part of the attributes missing and almost no products have all the attributes available.}
\label{fig:statusquo}
\end{figure}

\section{Predicting attributes}
\label{sec:attr}

To quote Niels Bohr, ``prediction is very difficult, especially if it is about the future''. In our case, it is rather about the past and the present (product data), but it remains a non-trivial affair. In this section, we present the challenges we face, the approaches we consider, and we give an overview of the current results.

\subsection{Motivation}
The fact that different divisions within the company produce and consume different product attributes presents particular challenges regarding the coverage and consistency of this information. 

Retail, for instance, uses specific ``personas'' to represent the core customer groups it targets. These are pre-defined characters that epitomise a particular fashion taste or style, often in association with particular brands. The retail persona, or customer segment, despite its name, is implemented as a product attribute: a dress may identify a `girly girl' persona, or rather a `glam girl', and so on. If this information, whenever applicable, were available for all the products customers interact with, that would allow for a better understanding of the customers' behaviour and preferences. It would provide actionable insight into how to improve customers' experience by providing the most relevant products to match the customers' unique taste. Unfortunately, if we take the category of dresses as a representative example, the customer segment is missing across 75\% of the catalogue history. Moreover, the segment's taxonomy changes over time in accordance with fashion trends.

Likewise, even more straightforward product aspects, such as the design pattern, are often missing: 70\% of all dresses carry an empty pattern label. That keeps us from being able to analyse or forecast the sales performance of `floral' dresses for instance or to explicitly recommend dresses with `animal' prints, etc. In addition, any model that requires multiple attributes at a time would have very few input samples to learn from, since, even within the same category, different products might be missing values for different attributes. For example, design pattern and customer segment are jointly available only on about 5\% of all dresses. Figure~\ref{fig:statusquo} provides an illustration of the situation, which essentially applies to all the product attributes that are not customer-facing.

In order to enable more in-depth analytics and better data products such as content-based and hybrid product recommendations --- which will be discussed in Section~\ref{sec:recs} --- it is then necessary to consolidate the available product data by filling in the missing attribute values. To this end, we implemented a multi-task multi-modal neural network to leverage all the unstructured information that is available on product pages: namely, product images, product name, brand, and description, as illustrated in Figure~\ref{fig:probleminputoutput}. Our methodology is detailed in the following sections.

\subsection{Design}
We illustrate next the design choices we made to be able to predict product attributes at scale from partially-labelled samples.

\subsubsection{Image Classification}
Fashion is a very visual domain and, with the popularity of Deep Learning, there already exist in the literature several approaches to classify and predict clothing attributes from images, both from models trained end-to-end or from intermediate representations: e.g.~\cite{inoue2017multi, liu2016deepfashion, schindlerfashion}. Our setup is peculiar as, along with images, we have textual information to leverage and we need to be able to do so at scale. Therefore, image processing becomes part of a bigger pipeline where visual features extracted from products' shots are precomputed and stored, in order to speed up the training of the attribute prediction model, but also in order for said features to be readily available for other applications. 

The visual feature generation step is an example of \emph{representation learning}; to this end, we apply a VGG16~\cite{simonyan2014very} Convolutional Neural Network with weights pre-trained on ImageNet. Although not any more state-of-the-art in classification, this architecture is still widely applied to the related tasks of object detection, image segmentation and retrieval due to the transferability of its convolutional features~\cite{zhai2017visual}.
For each product shot, we extract the $7\times 7 \times 512$ raw features from the last convolution layer before the fully-connected ones and perform a global max-pooling operation on those. This results in $512$ visual features for each product shot: each image is thereby projected onto a so-called \emph{embedding} space.

\subsubsection{Text Classification}
Convolutional neural networks (CNN) have proved to be effective at classifying not only images but also text~\cite{kim2014convolutional}. Indeed, sentences can be treated as word sequences, where each word in turn can be represented as a vector in a multidimensional \emph{word embedding} space. A 1-D convolution over this representation, with filters covering multiple word embeddings at a time, is an efficient way to take word order into consideration~\cite{johnson2014effective}. Following~\cite{kim2014convolutional}, we experimented with both fixed, i.e. pre-trained embeddings and free, randomly initialised ones; the latter performed significantly better in our case. We also made sure that the chosen architecture had an advantage over a well-tuned linear classifier over normalised bag-of-words (TF-IDF), which provides a strong baseline since words in the product description often contain ---or directly relate to--- the target attribute value.

The central branch of the diagram in Figure~\ref{fig:multi-nn} depicts our text processing pipeline: a simple 1D convolution over the word embedding layer, followed by a max-over-time pooling of convolutional features, and a fully-connected layer. In fact, since our product captions are often short (e.g., 50 words or less), we did not need any custom pooling layer despite the multi-label nature of our problem~\cite{liu2017deep}. In order to build enough capacity to be able to learn multiple attributes at a time, we just increased the number of filters and the number of neurons in the convolutional and in the dense layer, respectively~\cite{zhang2015sensitivity}.

\subsubsection{Multi-modal Fusion}
Merging the visual features of an image with the semantic information extracted from the text accompanying said image, has already proved helpful in object recognition tasks~\cite{frome2013devise}, especially when the number of object classes is large or fine-grained information is hard to gather from images alone. Regarding applications to fashion in particular, there exist multi-modal approaches to sales forecasting~\cite{bracher2016fashion}, product detection~\cite{rubio2017multi}, and search~\cite{laenen2017cross}.

\begin{figure}[t]
\includegraphics[width=\columnwidth]{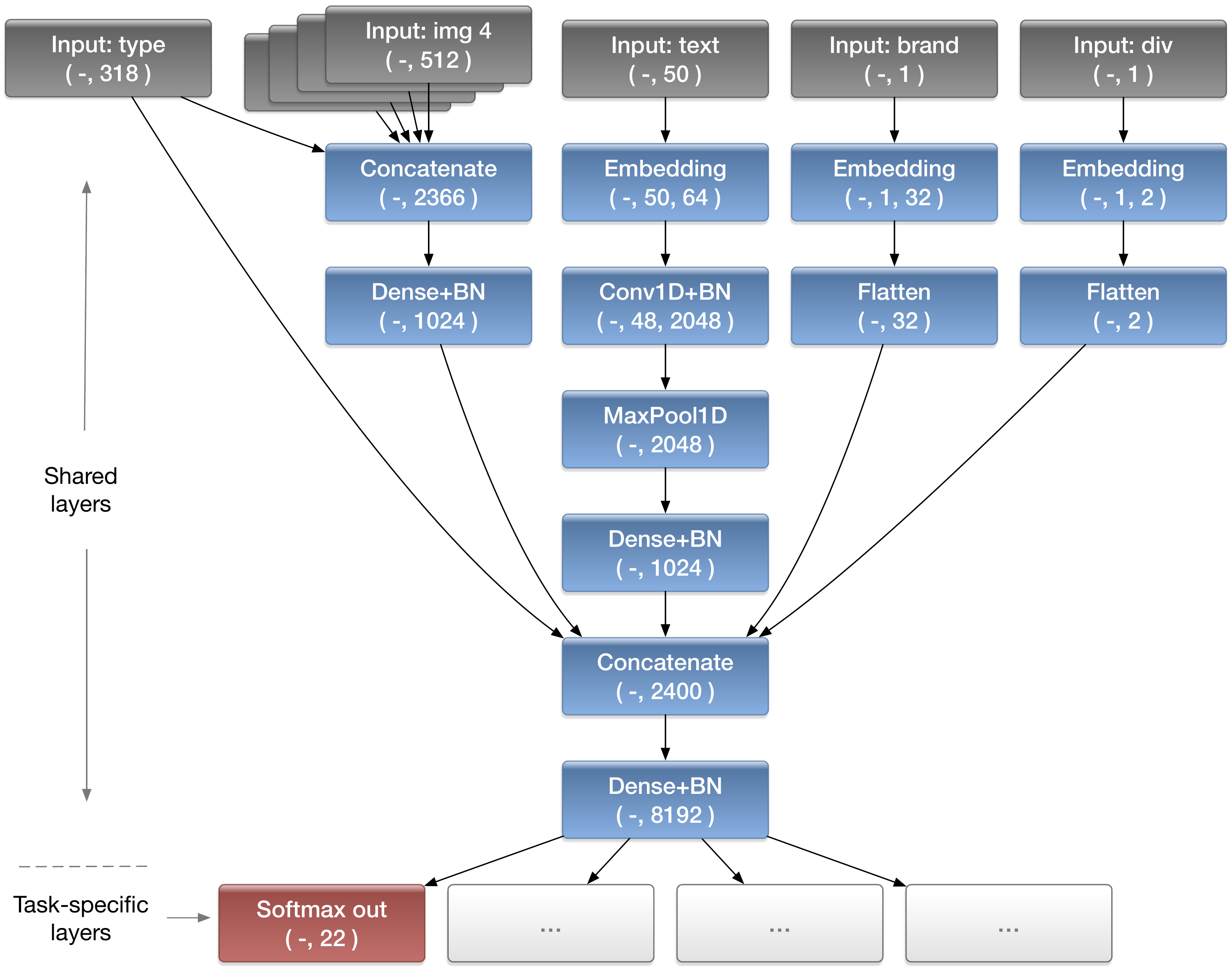}
\caption{Multi-modal multi-task architecture. In grey the inputs to the network, in blue the hidden layers and in red the output layers (one per task, i.e., attribute). In parenthesis, the output size of each layer, which reflects the input encoding for input layers, the embedding dimensionality for embedding layers, and the number of filters and neurons for convolutional and dense layers, respectively.}
\label{fig:multi-nn}
\end{figure}

Zahavy et al.~\cite{zahavy2016picture} solve a closely-related problem, product classification in a large e-commerce; they use an image CNN, a text CNN, and investigate possible fusion policies to merge the respective CNN outputs. Our solution, besides having to deal with partially annotated labels, as will be discussed in the next section, also consumes additional product metadata (product type, brand, and division), and merges the encodings from all these inputs via the architecture depicted in Figure~\ref{fig:multi-nn}. 
In particular, the product type information is used both when merging pre-computed image embeddings, since full-body shots contain multiple products, and when merging the visual features with the semantic information (product title and description) processed by the text CNN, along with brand and division embeddings that are trained end-to-end. Notably different from~\cite{zahavy2016picture}, is the structure of the output layers as discussed next.

\subsubsection{Multi-Task learning with Missing Labels}

The most straightforward solution to the multi-label scenario would be to one-hot encode and concatenate all target labels into a single output, then train a network with binary cross-entropy as a loss function and sigmoid activation in its last layer~\cite{zahavy2016picture,liu2017deep}. However, this approach would treat all classes across all labels as being mutually independent, whereas in reality each target attribute only takes a subset of all possible target values. In related work for multi-label image classification and annotation, state-of-the-art approaches aim to also model these labels dependencies, for instance by stacking a CNN with a recurrent layer (RNN) that processes the labels predicted by the CNN~\cite{wang2016cnn}.
Nevertheless, the prominence of missing labels in our training data would require implementing a custom loss function or masking layers in order to avoid propagating errors when a label is not available. For this reason, we opted for a simpler multi-task approach~\cite{ramsundar2015massively}, which still allows for a degree of label hierarchy because it matches each target attribute with a cross-entropy loss, whereby the predicted class is chosen amongst the possible values for that attribute only. That is, each target becomes a separate, imbalanced, multi-class problem. Hence, we can implement custom weighting schemes for the loss of the different targets depending on the labels frequencies of the specific attribute~\cite{li2017retrieving}.

A multi-task network provides us with an effective way to learn from all the products for which at least one of the target attributes has been labelled, thereby performing implicit data augmentation, and also implicit regularisation since simultaneously fitting multiple targets should favour a more general internal representation~\cite{ruder2017overview}. Conversely, even disregarding the added complexity of having to maintain multiple models in production, single-task networks for individual attributes could only leverage a fraction of the data (see Table~\ref{tab:test_attr}, second column), with a higher risk of overfitting. Even worse, as previously discussed, a single-task multi-label model would just not have enough training data without a clever way to deal with missing labels.

In practice, as illustrated by Figure~\ref{fig:multi-nn}, we build one model per attribute, but they all share the same parameters up to the output layer, following the `hard' parameter-sharing paradigm~\cite{ramsundar2015massively,ruder2017overview}. We also prepare one dataset per attribute. During training, in a random sequential order, we update each model with a gradient descent step on a randomly sampled mini-batch from its corresponding dataset. The hard sharing implies that each stochastic gradient descent (SGD) step also updates the common layers' parameters; a small learning rate and the randomised update order ensure that training is stable. 

As a matter of fact, preliminary experiments produced instability when using adaptive optimisers with momentum~\cite{kingma2014adam}; namely, validation set accuracy seemed to stall faster and to oscillate more across the various models. There is empirical evidence that adaptive methods might yield solutions that generalise worse than those from simple SGD with momentum~\cite{wilson2017marginal}. However, it is not clear whether that applies to a multi-task setting with parameter sharing. In our case, the simple addition of momentum to SGD seemed to already hurt training performance. Hence, we eventually resorted to using vanilla SGD with a constant learning rate and no momentum.


\subsection{Experimental Setup}

We collected data for all products ever available on any of our platforms up to September 2017, which had a product type, four image shots and a text description available, resulting in about 883K samples. The problem inputs and outputs are illustrated in Figure~\ref{fig:probleminputoutput}.
\begin{table}[t]
\caption{List of the attributes that we predict. For each attribute, the number of labelled samples, the number of classes, the weighted $F_1$ score on the test set (weighted by the support of each label), and the test set accuracy are given. The last column provides the frequency of the majority class in the test set (baseline accuracy), indicative of the accuracy of a naive model.}
\label{tab:test_attr}
\centering
\resizebox{0.48\textwidth}{!}{%
\begin{tabular}{lccccc}
\toprule
attribute &\#samples &\#classes &  \minibox[c]{weighted \\ $F_1$} &accuracy & \minibox[c]{baseline\\ accuracy}\\
\midrule
selling period &771.1K &33 &0.361 &0.361 &0.074\\
price range &717.0K &3 &0.948 &0.946 &0.843\\
style &543.8K &186 &0.847 &0.846 &0.072\\
collection &353.0K &6 &0.999 &0.999 &0.878\\
buying period &290.7K &3 &0.928 &0.928 &0.778\\
sleeve length &260.3K &11 &0.843 &0.840 &0.348\\
exclusive &259.9K &2 &0.958 &0.958 &0.864\\
back to school/college &237.8K &2 &0.946 &0.946 &0.922\\
customer segment &230.8K &19 &0.807 &0.807 &0.178\\
pattern &224.7K &22 &0.829 &0.828 &0.460\\
leather/non leather &154.5K &2 &0.950 &0.949 &0.713\\
fit &130.5K &20 &0.906 &0.907 &0.460\\
neckline &129.3K &16 &0.865 &0.868 &0.495\\
product fit &111.8K &13 &0.930 &0.931 &0.643\\
dress length &109.3K &3 &0.908 &0.907 &0.625\\
seasonal event &109.1K &6 &0.981 &0.981 &0.624\\
womenswear category &102.6K &56 &0.823 &0.823 &0.084\\
gifting &96.9K &2 &0.983 &0.983 &0.964\\
product length &87.7K &6 &0.935 &0.937 &0.775\\
dress type &86.2K &5 &0.675 &0.677 &0.383\\
shirt style &72.2K &3 &0.966 &0.967 &0.723\\
use/occasion &55.3K &13 &0.665 &0.666 &0.238\\
jewellery finish &28.6K &11 &0.664 &0.678 &0.378\\
jewellery type &24.5K &23 &0.821 &0.823 &0.117\\
denim wash colour &17.4K &9 &0.631 &0.632 &0.290\\
shirt type &14.3K &5 &0.911 &0.912 &0.488\\
heel height &13.1K &4 &0.896 &0.894 &0.590\\
denim rip &12.8K &5 &0.922 &0.914 &0.531\\
accessories type &9.6K &6 &0.985 &0.985 &0.296\\
weight &7.5K &3 &0.845 &0.843 &0.558\\
bags/purses size &3.3K &5 &0.879 &0.887 &0.265\\
high summer &2.7K &2 &0.979 &0.979 &0.726\\
suit type &2.2K &4 &0.730 &0.742 &0.505\\
cold weather &0.8K &2 &0.752 &0.754 &0.739\\
\bottomrule
\end{tabular}
}
\end{table}
For each of these products, we also collected all manually annotated attributes that contained missing values, as illustrated in Figure~\ref{fig:statusquo}. 
Although not all attributes might apply to all product types, coverage can be low (see Table~\ref{tab:test_attr}, column \#samples).
The manual annotations in historical data do not always come with a fixed taxonomy, are not always curated and in some cases might contain incorrect labels. To construct a ground-truth, we use a set of heuristics to determine which attributes apply to which product types as well as specify a minimum number of samples for a possible value of an attribute to be predicted, which we set to 500.

We empirically tuned the hyper-parameters of the architecture for a balance between computational cost and performance. For the evaluation, we took 90\% of the data as train and 10\% as test. We use as mini-batch size 64 with a learning rate of 0.01. Our implementation uses Keras~\cite{chollet2015keras} with a Tensorflow~\cite{abadi2016tensorflow} backend.

The evaluation task is to predict all the attributes that are applicable. 
We report the weighted $F_1$ score, i.e. for each task we average the individual $F_1$ scores calculated for each possible label value, weighted by the number of true instances for each label value.

\subsection{Experimental Evaluation}

\begin{figure}
\subcaptionbox{Average (line) and 0.95 CI (shade) across all attributes.}
	{\includegraphics[width=0.48\textwidth]{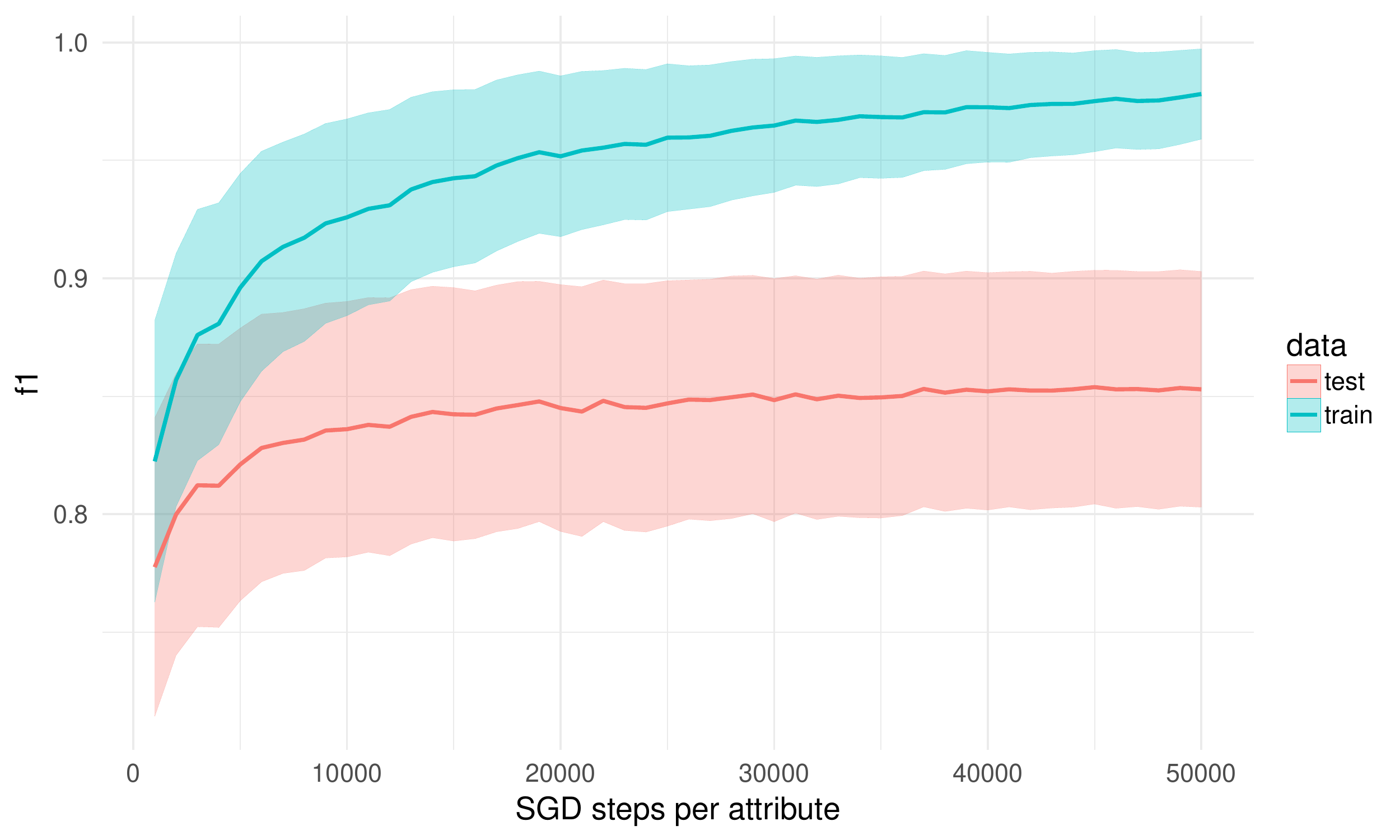}}
\subcaptionbox{Test set performance for each attribute. 
}{\includegraphics[width=0.48\textwidth]{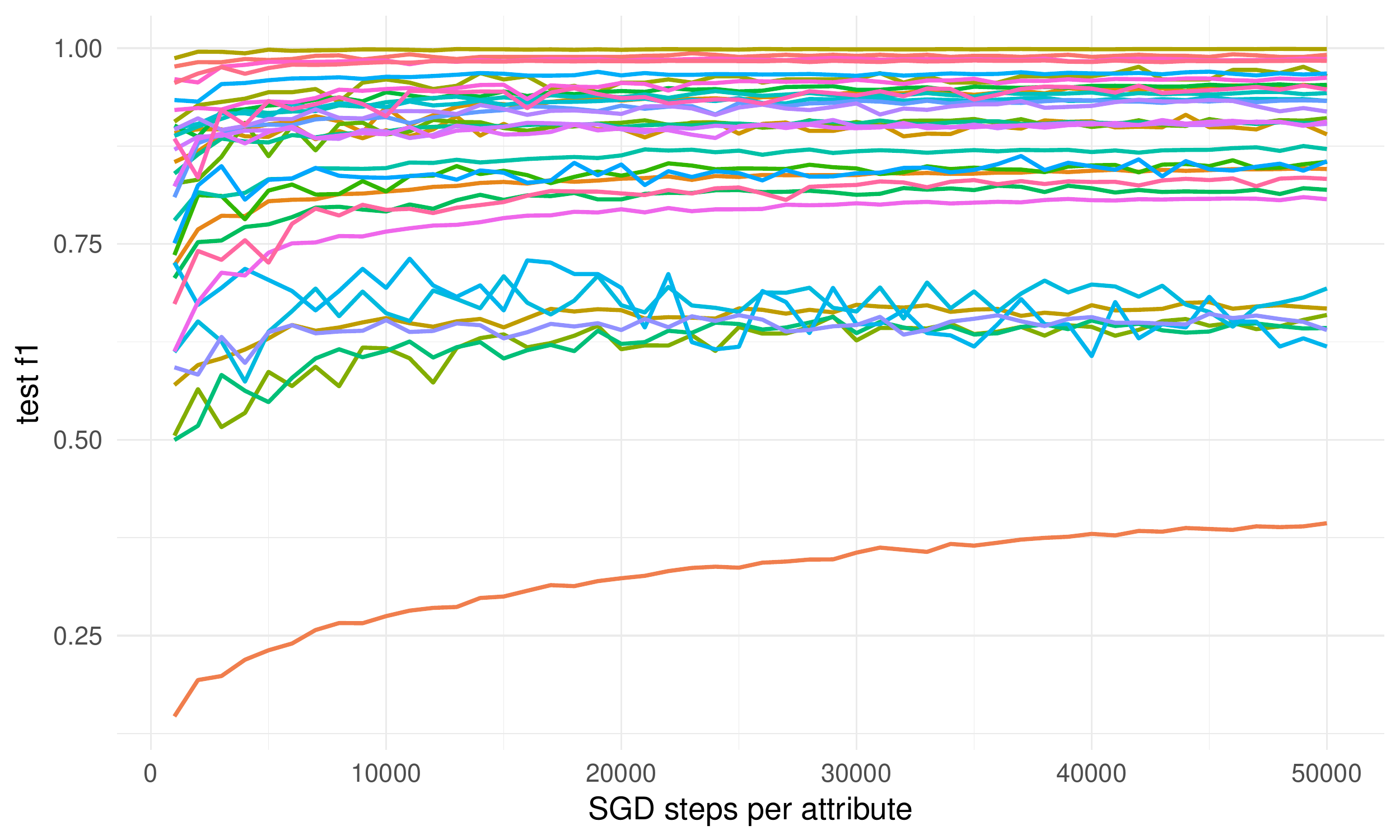}}
\caption{Evolution of the weighted $F_1$ scores, measured at intervals of 1000 steps per attribute (minibatch size of 64). The model achieves an average weighted f1 of 0.855. The score is still increasing in train but plateaus out in test.}
\label{fig:trainevol}
\end{figure}

\begin{figure}
\centering

\subcaptionbox{Neckline, weighted $F_1=0.865$\label{fig:neckline}}{%
	\includegraphics[width=0.48\textwidth]{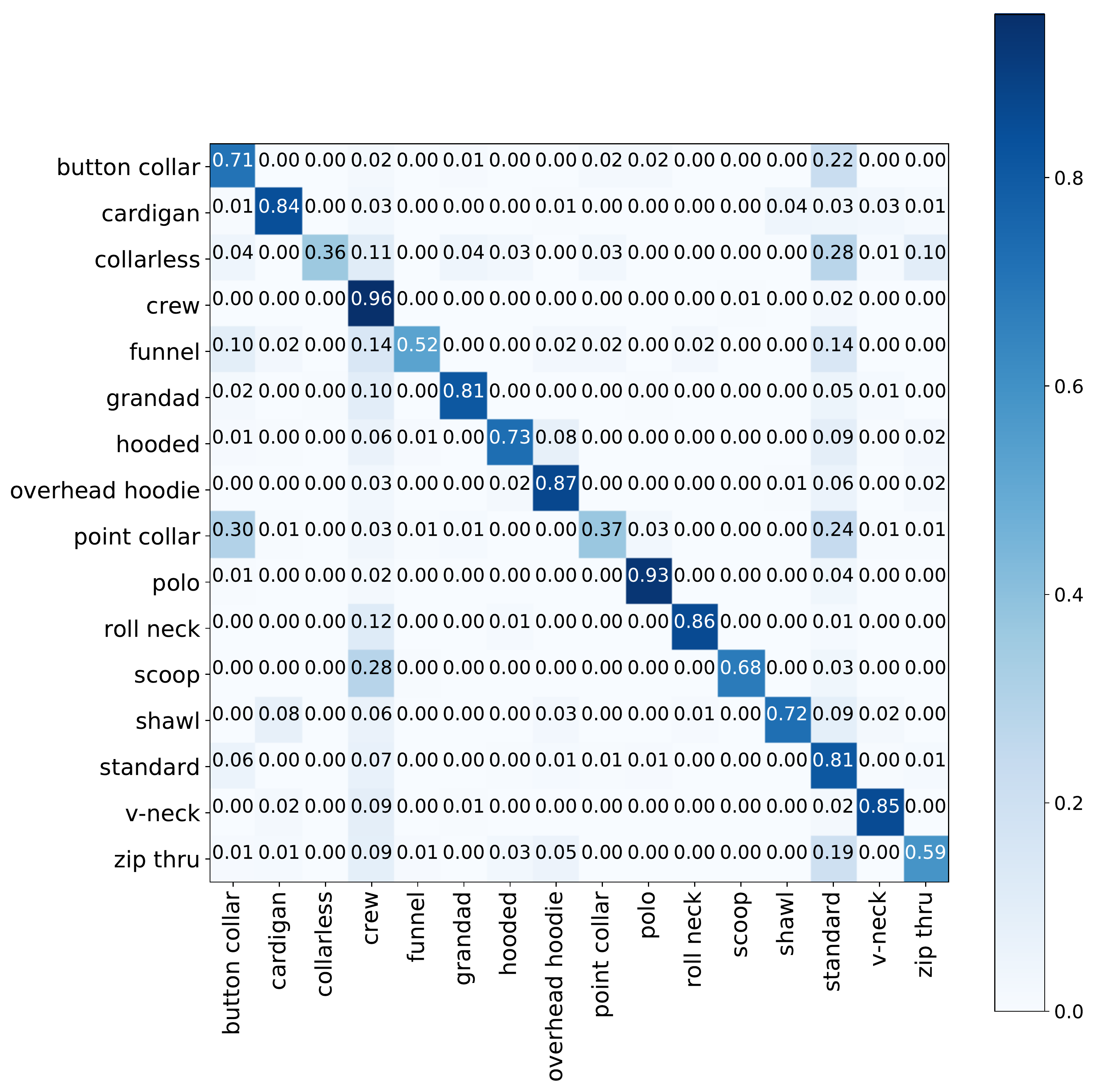}}
\subcaptionbox{Use/Occasion, weighted $F_1=0.665$\label{fig:use}}{%
	\includegraphics[width=0.42\textwidth]
    {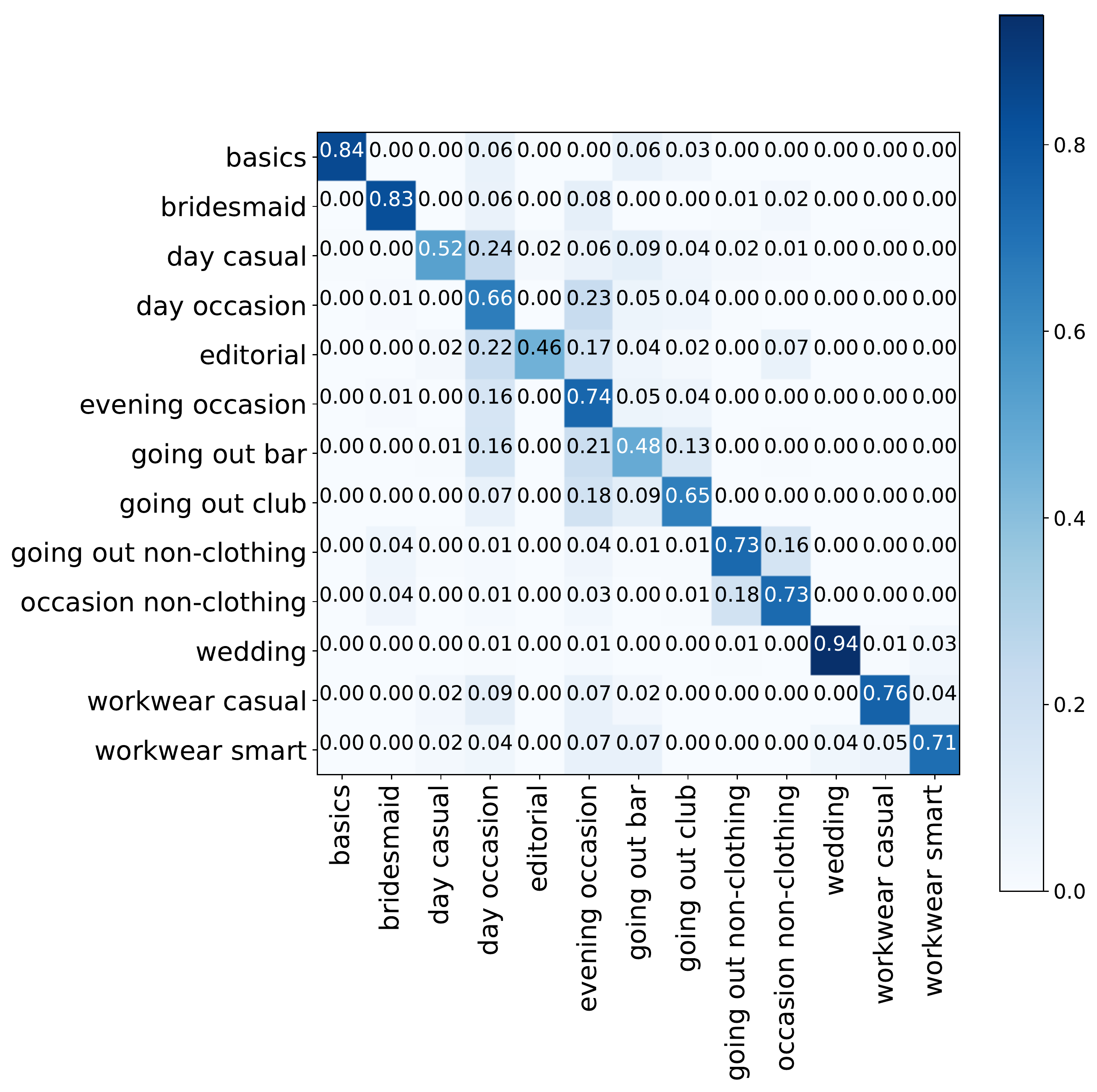}}

\caption{Examples of test-set normalised confusion matrices for target attributes; true labels (by row) vs predicted labels (by column). Attributes might be related to a product shape or other visual characteristics (a), but also to higher-level properties that are not directly expressed in the product data (b). Regarding the former, although the model does well overall, one can still observe biases towards the most popular classes such as `crew' neck and `standard' neck. Regarding the latter, the easiest end use to predict is `wedding' whereas the most challenging is `going out bar', which is often confused with `evening occasion'.}
\label{fig:confmatrix}
\end{figure}
\begin{figure}
\includegraphics[width=0.48\textwidth]{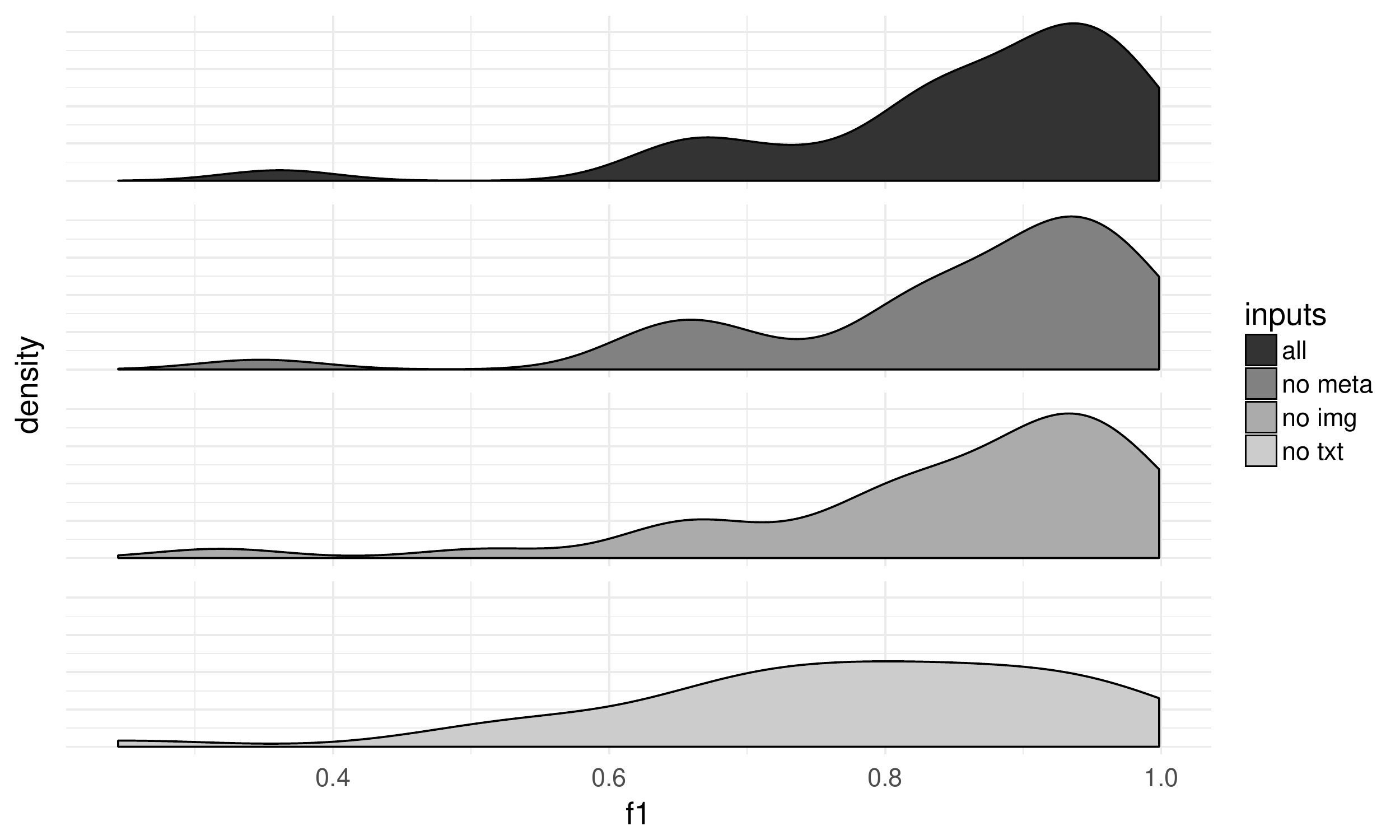}
\caption{Ablation study. Density distribution (y-axis) of test $F_1$ scores (x-axis) across all predicted attributes when removing one of the inputs in turns (see legend). The average $F_1$ scores are: 0.855 for the full model, 0.849 without product metadata (product type, brand, and division), then 0.842 without image features, and 0.773 without text input.}
\label{fig:ablation-dist}
\end{figure}

The results of this evaluation are summarized in Table~\ref{tab:test_attr}. We can see that the model always outperforms the naive baseline (majority class) in terms of accuracy. The evolution of performance during training is shown in Figure~\ref{fig:trainevol}, the model performance in test is plateaus after 50K optimisation steps (mini-batches of 64 samples) per attribute.
The confusion matrices for two of the attributes are shown in Figure~\ref{fig:confmatrix}. We can see that for neckline the mass is heavily concentrated on the diagonal. On the off-diagonal, the most significant mismatch is between `button collar' and `point collar', both types of shirts neckline. Samples from the class which has the least support (`point collar'), are often misclassified  as `button collar', which is five times more frequent. The use/occasion attribute represents a higher-level categorization that is harder to capture from the data; still, the performance is way above chance level with a 0.665 accuracy (vs 0.238 baseline, 13-class task), with `wedding' being the easiest class to spot.
The average of the weighted $F_1$ scores is 0.855 when using all inputs (image, text, metadata). The most important input is text: without it, the $F_1$ score drops to 0.773. Images are the second most important: without them, the $F_1$ score drops to 0.842. Finally, metadata is the least important: without it, the score drops to 0.849. The relative importance of image and text is in line with what has been observed in a related problem and modelling approach~\cite{zahavy2016picture}, where a text CNN alone can classify e-commerce products better than an image CNN alone, and multi-modal fusion only brings a marginal improvement. The impact of removing each input can be seen in more detail in Figure~\ref{fig:ablation-dist}.

\section{Application to personalised recommendations}
\label{sec:recs}

We illustrate next one of the applications of our attribute extraction pipeline, improving personalised product recommendations. First, we start by motivating the need for a content-aware recommendation solution. Then, we describe the design of our recommendation algorithm and how it includes information about the content of the products. Finally, we provide some experimental evidence of the effectiveness of our approach.

\subsection{Motivation}

Recommender systems are one of the most common approaches for helping customers discover products in vast e-commerce catalogues. In ASOS, we rely on personalised recommendations to provide a source of inspiration for our customers and to anticipate their needs. In established e-commerce sites with an abundance of data, collaborative filtering has proven to be a good starting approach for delivering accurate, personalised recommendations to customers~\cite{Linden2003}. Collaborative filtering algorithms produce recommendations for customers based on other customers' interactions, either by finding direct relationships between customers and between products ---e.g.\  products bought by the same customers tend to be similar---, or by finding latent associations between customers and products based on their interactions ---e.g.\ using dimensionality reduction techniques. Despite their popularity, it is also well known that one of the main weaknesses of collaborative filtering is the so-called \textit{cold-start problem}, that is, the fact that these type of algorithms tend to perform poorly for customers and products with little (or no) interaction data. Content-based algorithms, on the contrary, do not suffer from a product cold-start situation but, overall, tend to perform worse than their collaborative filtering-based counterparts. Furthermore, they depend importantly on consistent and structured product descriptions and metadata.

In ASOS, a dynamic product catalogue and the ability to offer customers the latest fashion trends are key aspects of the business. This indicates that neither collaborative filtering ---which suffers from product cold-start, notably in the case of newly-added products--- nor content-based approaches ---often suffering from low accuracy and sensitive to data quality--- are, in isolation, sufficient for providing customers with an optimal discovery experience. As a solution, we propose the use of a hybrid recommendation approach, which combines both approaches and overcomes their respective limitations. Moreover, our solution for extracting product attributes contributes to the success of content-aware approach by ensuring the quality of the data that is used to compute recommendations.

\subsection{Design}

Our proposed approach falls within the family of \textit{factorisation models}, matrix factorisation being the basic, best-known approach to them. In essence, factorisation models try to represent both customers and products in a high-dimensional vector space ---in which the number of dimensions is orders of magnitude smaller than the number of either customers and products--- where relevant, latent properties of customers and products are encoded. Vectors are typically optimised so that dot products between customer vectors and a product vectors define meaningful scores that help determine rankings or top-N selection of products to be displayed to individual customers.

\begin{figure}
\includegraphics[width=0.42\textwidth]{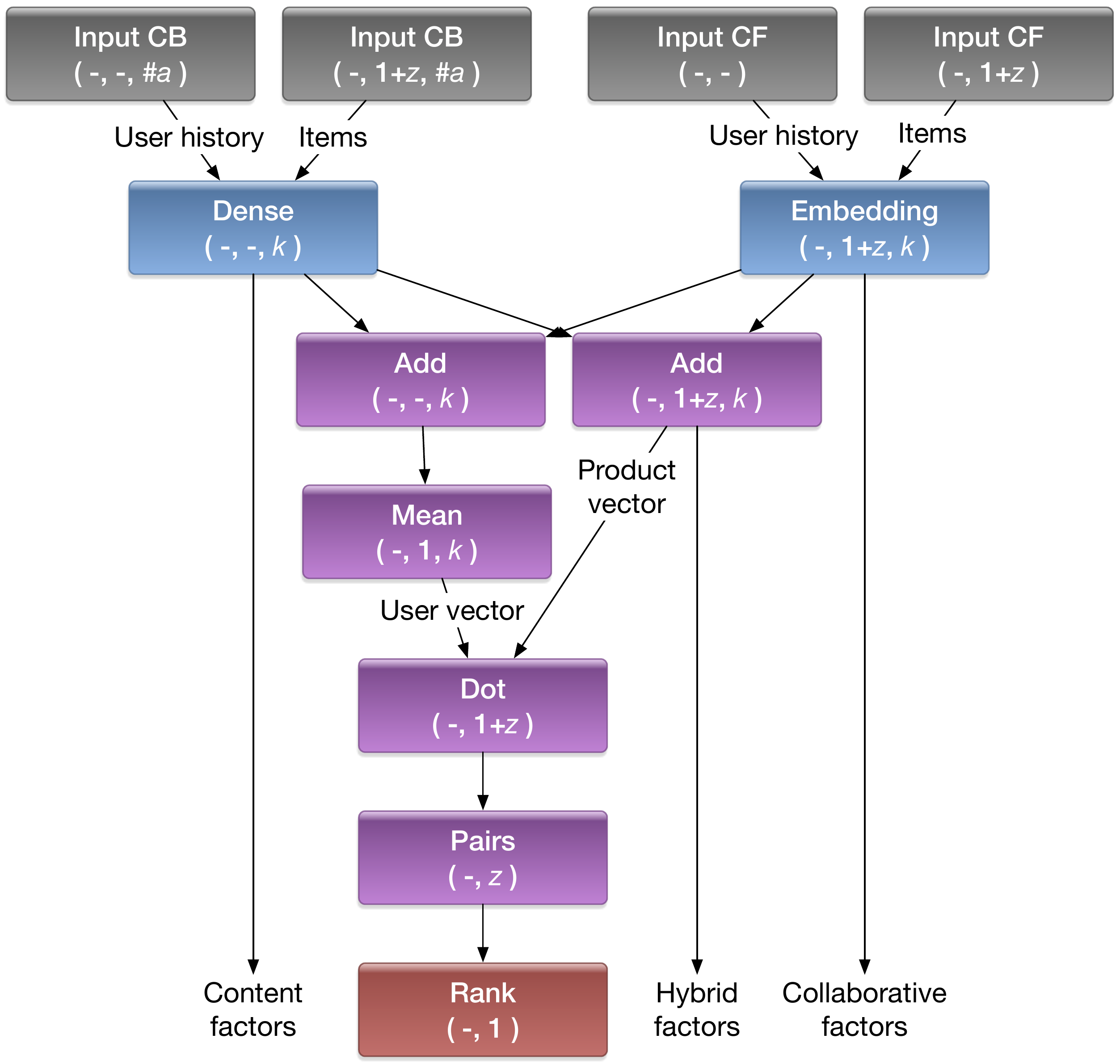}
\caption{Hybrid recommender architecture. Content-based (CB) and collaborative (CF) inputs to the network are depicted in grey, hidden layers in blue, functional layers in purple, and final layer in red. In parenthesis, the output size of each layer: $\#a$ denotes the dimensionality of the encoded attributes, $z$ the number of negative samples per batch in the ranking loss, and $k$ the dimension of the product factors.}
\label{fig:krecs}
\end{figure}

In a collaborative filtering setup, which is the case of most classic matrix factorisation approaches, customer and product vectors have to be learned as parameters of the model. In neural networks terminology, this means that our recommendation model includes two embedding layers, one for customers and one for products. Nevertheless, we go beyond this configuration and extend it for our purposes. For example, instead of embedding layers for products, we also rely on a content-based approach and use features associated with products (attributes, brand, etc.) to produce said product vectors. The mapping between product features and vectors can be achieved by, e.g., a feed-forward neural network. Moreover, it is possible to combine vectors coming from an embedding layer with those coming from a feature-fed neural network, thus producing a \textit{hybrid} representation of products. This forms the basis of our hybrid recommendation algorithm, where we have found out that embedded and content-derived vectors can be combined by summing them~\cite{Kula2015}. In particular, our approach leverages product type, price, recency, popularity, text descriptions ($\approx$1K), image-based features ($\approx$0.5K) and, importantly, predicted product attributes ($\approx$0.5K). The content-based component takes as input \#a ($\approx$2.4K) attributes per product.

A second modification that we perform with respect to classic matrix factorisation is the so-called \textit{asymmetric matrix factorisation}~\cite{Steck2015}. This method aims to reduce the complexity of the model by eliminating the need to learn an embedding layer for customers. Instead, vectors for customers $v_u$ are computed as an intermediate representation. Specifically, we do this by gathering the vectors $v_i$ of the set $I_u$ of products the customer $u$ has interacted with, and by combining them, e.g.\ simply by taking the average: $v_u = \text{avg}_{i \in I_u} v_i$. Note that, in cases where customer attributes are to be taken into account (gender, country, estimated budget), it is possible to combine customer attributes with product vectors using a neural network, as in the approach of Covington et al.~\cite{Covington2016} or, as proposed by Rendle~\cite{Rendle2010} or Kula~\cite{Kula2015}, by doing vector composition. Notably, eliminating the customer embedding layer from the model reduces drastically the size of the model, which has positive consequences in terms of improving the scalability of the approach and reducing the risk of over-fitting.

Finally, we provide some details about the optimisation criterion that we are using to train the model. Following the recent stream of learning to rank approaches~\cite{Karatzoglou2013}, we minimise during training a loss function that models a ranking problem where positive interactions in training ---such as products that have been purchased by a customer in our training data--- have to be ranked higher than a sampled selection of negative items (items that are believed not to be relevant to the customer based on lack of interaction). In particular, we have chosen the WMRB loss function~\cite{Liu2017}, which is a batch-oriented adaptation of the WARP loss function~\cite{Weston2011}, which has proven to be effective at producing good item rankings. This function, given a customer $u$, a \textit{positive} product $i$ and a random selection $Z$ of \textit{negatively sampled} products, tries to find appropriate scores $s_{u,i}$ from the factorisation model so that the rank of positive item $i$ when ordered with the products in $Z$ by decreasing score is as low as possible. Given that the exact rank is not differentiable, WMRB optimises the following approximation instead:
\begin{equation}
L_{WMRB}(u, i, Z) = \log \sum_{j \in Z} \left| 1 - s_{u,i} + s_{u, j} \right|_{+}
\end{equation}
where $\left| \cdot \right|_{+}$ is a rectifier function. The resulting architecture design is summarised in Figure~\ref{fig:krecs}.

\begin{figure*}[!h]
\centering
\begin{tabular}{c c@{}c@{}c c@{}c@{}c c@{}c@{}c}
\toprule
Query &\multicolumn{3}{c}{Collaborative} &\multicolumn{3}{c}{Content} &\multicolumn{3}{c}{Hybrid}\\
\midrule
\includegraphics[width=0.09\textwidth]{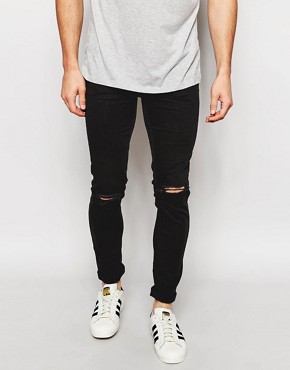} &\includegraphics[width=0.09\textwidth]{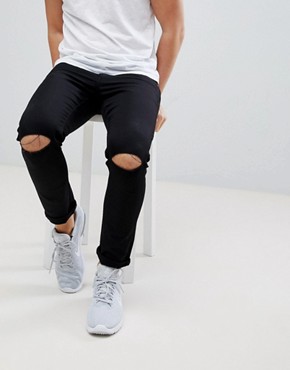} &\includegraphics[width=0.09\textwidth]{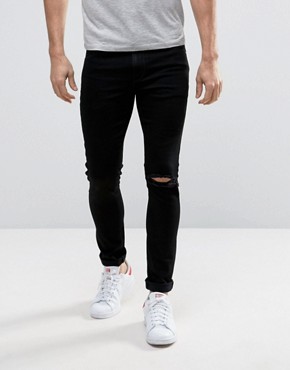} &\includegraphics[width=0.09\textwidth]{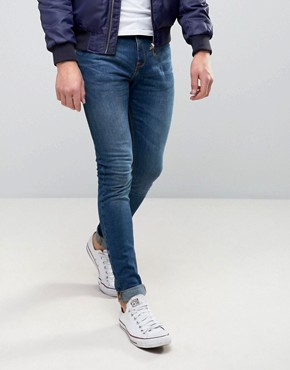} &\includegraphics[width=0.09\textwidth]{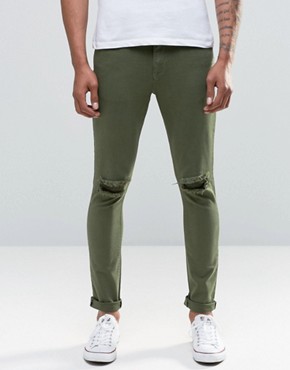} &\includegraphics[width=0.09\textwidth]{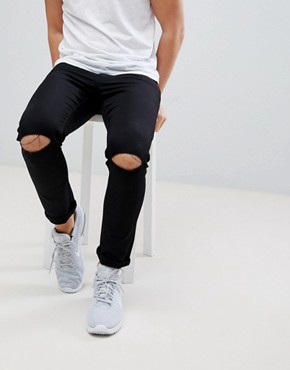} &\includegraphics[width=0.09\textwidth]{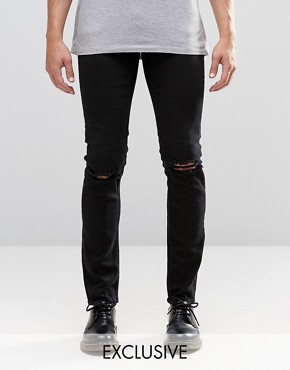} &\includegraphics[width=0.09\textwidth]{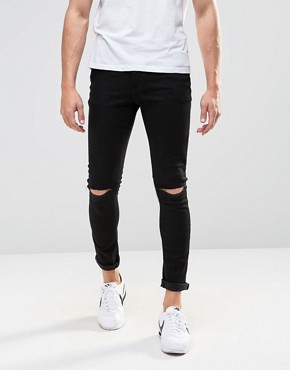} &\includegraphics[width=0.09\textwidth]{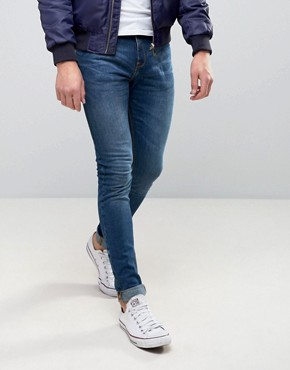} &\includegraphics[width=0.09\textwidth]{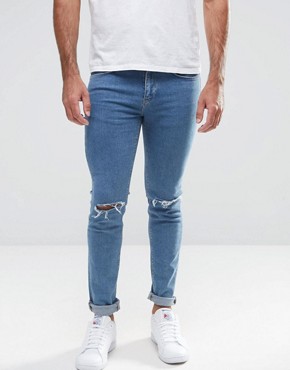}\\
\midrule
\includegraphics[width=0.09\textwidth]{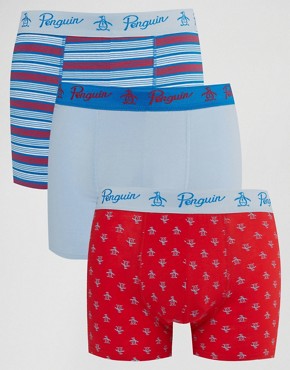} &\includegraphics[width=0.09\textwidth]{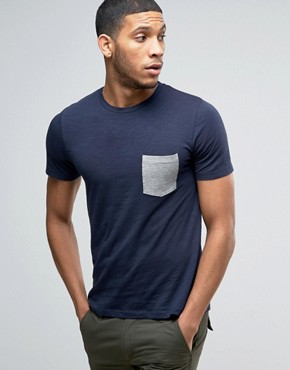} &\includegraphics[width=0.09\textwidth]{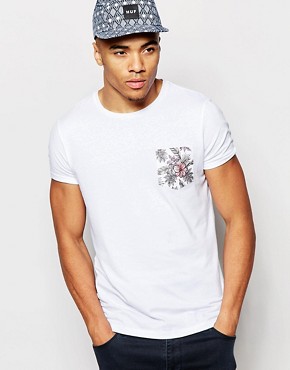} &\includegraphics[width=0.09\textwidth]{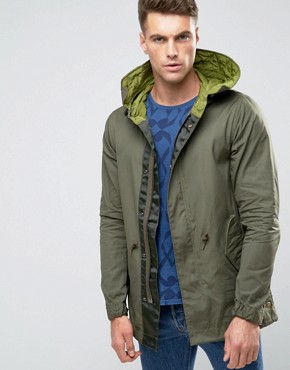} &\includegraphics[width=0.09\textwidth]{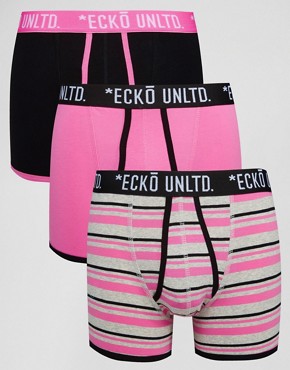} &\includegraphics[width=0.09\textwidth]{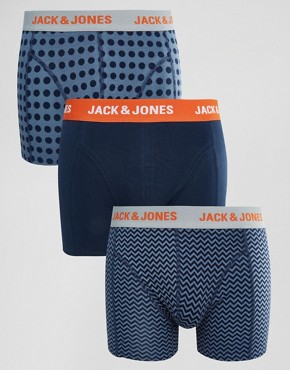} &\includegraphics[width=0.09\textwidth]{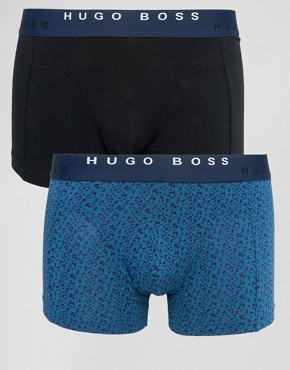} &\includegraphics[width=0.09\textwidth]{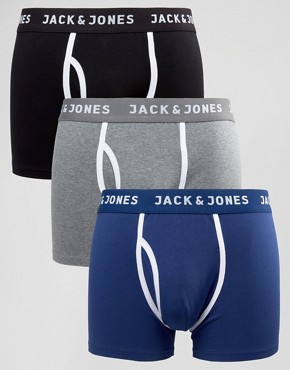} &\includegraphics[width=0.09\textwidth]{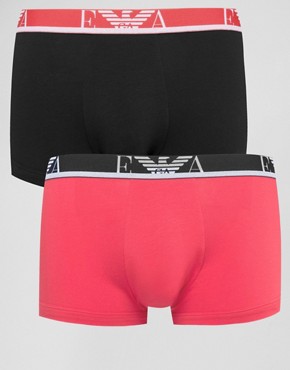} &\includegraphics[width=0.09\textwidth]{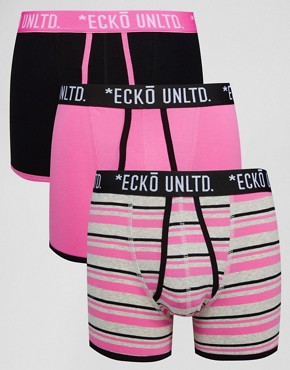}\\
\midrule
\includegraphics[width=0.09\textwidth]{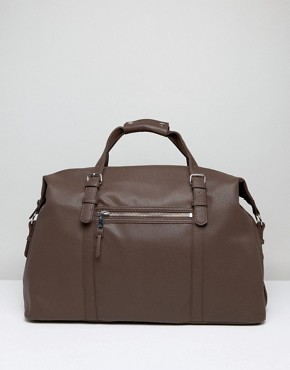} &\includegraphics[width=0.09\textwidth]{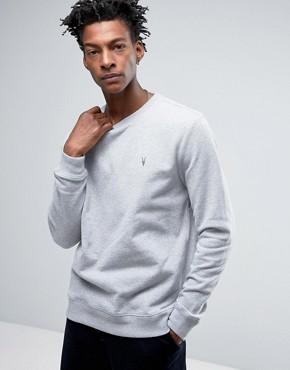} &\includegraphics[width=0.09\textwidth]{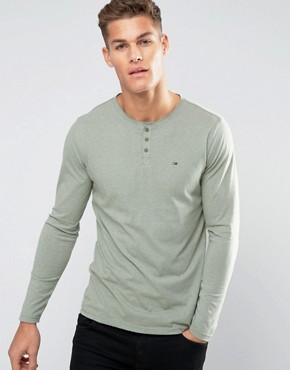} &\includegraphics[width=0.09\textwidth]{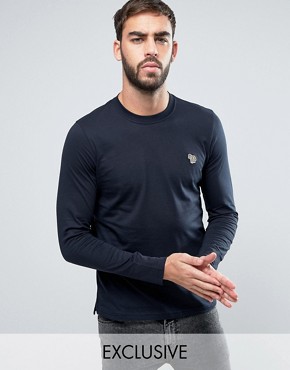} &\includegraphics[width=0.09\textwidth]{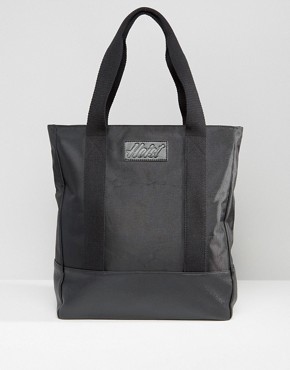} &\includegraphics[width=0.09\textwidth]{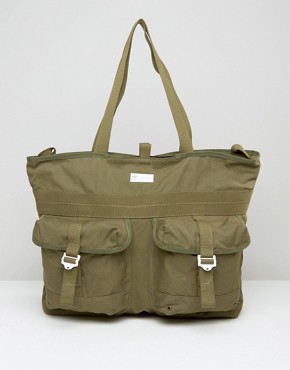} &\includegraphics[width=0.09\textwidth]{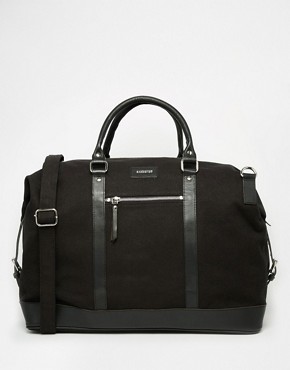} &\includegraphics[width=0.09\textwidth]{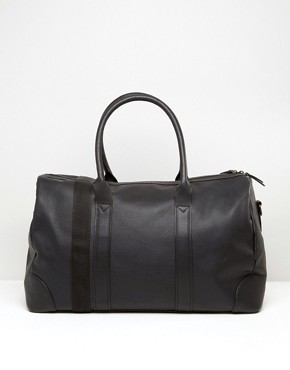} &\includegraphics[width=0.09\textwidth]{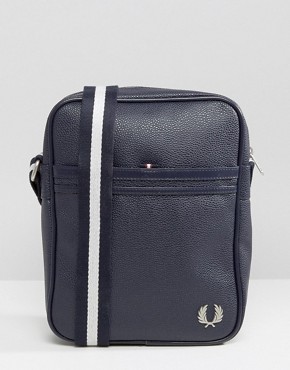} &\includegraphics[width=0.09\textwidth]{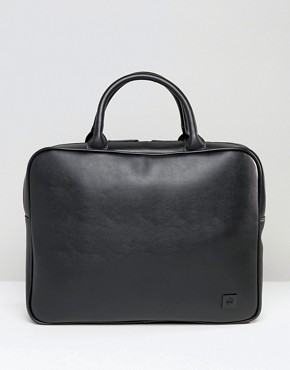}\\
\bottomrule
\end{tabular}
\caption{Examples of similar products defined by the recommendation models. Top row: product with the largest number of customer interactions in the training set. Middle row: product with a median number of customer interactions. Bottom row: least popular product in the training period. We can see that collaborative performs better for products with more interactions, while content and hybrid are not affected by this factor.}
\label{fig:recscomp}
\end{figure*}
\subsection{Experimental Setup}
In order to show the effectiveness of our proposed hybrid design, we have performed an experiment with a sample of our data. In particular, we have collected 13 months of interactions of our male customers starting from March 2016. The first 12 months have been used to train and validate the recommendation algorithms, whereas the last month was used for evaluation purposes. Out of all interactions, we kept only those that show a strong intention of buying (save for later, add to bag) together with actual purchases. Then, we randomly sampled 200K customers, ensuring that half of them had interactions in both training and test. In total, this amounted to 9M interactions with 117K products in the training set and 1M interactions with 54K products in the test set. We normalise all product features as follows: we first normalise to unit-norm each feature independently, and then we normalise to unit-norm each sample (i.e. each product).

The training procedure using the first 12 months of data goes as follows. Interactions of the last month of the training window (February 2017) were taken as the positive examples for the WMRB function, while negative instances were randomly sampled from the non-observed interactions with products that were available during the whole training window. As inputs to the predictions, i.e.\ interactions with products (and their features) to produce customer vectors, we chose for every customer 5 randomly sample interactions preceding the positive one whose rank we want to optimise in the loss function. This means that customers with less than 6 interactions (5 for input, 1 for prediction) were discarded from training. 

The evaluation task is to predict the purchases that customers made in the test period (March 2017). Note that this type of evaluation provides a lower bound of the performance of a recommendation, as it assumes that recommended products that were not purchased were not interesting to the customer. Another particularity of our task is that it freezes the learned model and its representations of the customer before the test period. This is helpful to illustrate the ability of the hybrid solution in dealing with cold-start products, i.e.\ those that are added to the catalogue immediately before or during the test period. In reality, we update our models daily so that we can alleviate the cold-start as much as possible. Precision and recall at cut-off 10 have then been used as the metrics to assess this task. 

To provide meaningful comparisons when discussing the validity of our hybrid approach, two baselines have been included using the same training and evaluation approach. First, we compare our approach with the obvious baseline for ranking tasks, i.e.\ popularity-based ranking~\cite{Amatriain2013}. Second, in order to see the benefit of fusing collaborative filtering and content, we also compared our hybrid design with collaborative-only and content-only versions. For all variations of our approach, we used the same vector size (k=200), the number of negative items $z = \left| Z \right| = 100$, batch size of 1024, and a variable number of epochs up to 100 that is determined by a validation set of 10\% of customers in training.

\vspace{-0.5em}
\subsection{Experimental Evaluation}
The results of this evaluation, summarised as the average values for each metric and algorithm across all customers in the test set, can be seen in Table~\ref{tab:baseline}. The first noticeable result, given by the low numbers for precision and recall, is that predicting customer activity days in advance is a difficult task. The results however, allow for the comparison between the different approaches. For instance, we can see that a pure content-based approach offers results that are only slightly better than what a non-personalised popularity-based ranking offers, which confirms our previous assumptions about the weakness of this approach. When looking at the pure collaborative filtering based solution, we can see that this approach is  superior to the baseline and also to the content-based one.  Finally, we observed that the hybrid model is the most effective of the compared approaches, as it outperforms the collaborative filtering approach, showing how bringing content into the model has a clear positive effect when delivering fashion recommendations.

\begin{table}[b]
\caption{Results of the recommendation experiments, averages and standard deviations over 10 runs. Best result on average in bold. Scores of prec@10 are statistically different between all methods at the 5\% significance level, differences in recall@10 are not for any of the methods.}
\label{tab:baseline}
\begin{center}
\begin{tabular}{lcc}
\toprule
& prec@10 & recall@10 \\
\midrule
Popularity & 0.00231 & 0.00765 \\
Collaborative & 0.00277$\pm$0.00011 & 0.00931$\pm$0.00036  \\
Content & 0.00246$\pm$0.00011 & 0.00755$\pm$0.00191\\
Hybrid & \textbf{0.00313$\pm$0.00015} & \textbf{0.00960$\pm$0.00252}\\
\bottomrule
\end{tabular}
\end{center}
\end{table}
We complement the quantitative evaluation just presented with a qualitative one, in which we visually analyse product similarities produced by the three versions of our designed recommender. The results can be found in Figure~\ref{fig:recscomp}. We select three seed products based on the number of customer interactions: the product with the highest number of interactions, a product with a median one and a product with the lowest value in the training set. This selection allows us to explore the trade-offs between collaborative filtering and content-based approaches when the number of interactions with products varies. Then we retrieved, for each seed product and algorithm, the top three products whose vectors have the highest cosine similarity with respect to the seed product's vector. The results offer some interesting insights. On the one hand, we can see that the collaborative filtering approach produces less obvious ---but not necessarily bad--- recommendations, especially when the number of interactions is low (middle and bottom row). On the other hand, we can see that the content-based approach defines similarities that are more explainable. By combining both methods, the hybrid recommender appears to benefit from the interpretability of the content, while producing recommendations that are different from ---and better than, as seen before--- the isolated approaches.

\section{Discussion and future work}
\label{sec:discussion}
Any business needs to understand its customers to be able to better serve them; in the digital domain, this revolves around leveraging data to tailor the digital experience to each customer. 
We have presented a body of work, which is currently being productionised, that enables such personalisation by predicting the characteristics of our products.  This is a foundational capability that can be used, for instance, to provide more relevant recommendations, as shown with the hybrid approach proposed. 

Regarding the attribute prediction pipeline, we plan to enhance the quality of the image features by first detecting which area of the picture contains the relevant product, since most fashion model shots contain multiple products. Another possible refinement is to fine-tune the image network ---now pre-trained on ImageNet--- on our own data, or to train the whole model end-to-end at an even higher computational cost. Regarding the application of the hybrid approach to other datasets, we expect the relative improvement of hybrid over collaborative filtering to grow with the sparsity of the user-item interaction matrix and to depend on the relevance of the product features. The optimisation of the hybrid model might also prove more challenging and, unlike with our dataset, require the regularisation of the activation/weights of the content/collaborative components.

As further future work we plan to use the capability to characterize products to enable additional use cases such as:
data-assisted design; better product analytics; sales forecasting; range planning; finding similar products; and improving search with predicted meta-data. Regarding the hybrid recommender system approach, we plan to study the trade-offs between content and collaborative in regimes ranging from cold-start to highly popular products.


\balance

\bibliographystyle{ACM-Reference-Format}
\bibliography{main} 

\end{document}